\documentclass[runningheads]{llncs}

\usepackage{eccv}

\usepackage{eccvabbrv}

\usepackage{graphicx}
\usepackage{booktabs}
\usepackage{microtype}
\usepackage{multirow}
\usepackage{array}
\usepackage{enumitem}
\usepackage{bbding}
\usepackage{bbm}
\usepackage[accsupp]{axessibility}  

\usepackage[pagebackref,breaklinks,colorlinks,citecolor=eccvblue]{hyperref}

\usepackage{orcidlink}

\begin{document}

\title{Boosting Semi-Supervised Temporal Action Localization by Learning from Non-Target Classes} 

\titlerunning{Learning from Non-Target Classes for SS-TAL}

\author{Kun Xia\inst{1} \and
Le Wang\inst{1} \and
Sanping Zhou\inst{1} \and Gang Hua\inst{2} \and Wei Tang\inst{3}}
\authorrunning{K. Xia, L. Wang, S. Zhou, G. Hua, W. Tang.}

\institute{
Institute of Artificial Intelligence and Robotics, Xi'an Jiaotong University \and
Wormpex AI Research \and University of Illinois Chicago\\
}

\maketitle

\begin{abstract}
  The crux of semi-supervised temporal action localization (SS-TAL) lies in excavating valuable information from abundant unlabeled videos. However, current approaches predominantly focus on building models that are robust to the error-prone target class (\ie, the predicted class with the highest confidence) while ignoring informative semantics within non-target classes. This paper approaches SS-TAL from a novel perspective by advocating for learning from non-target classes, transcending the conventional focus solely on the target class. The proposed approach involves partitioning the label space of the predicted class distribution into distinct subspaces: target class, positive classes, negative classes, and ambiguous classes, aiming to mine both positive and negative semantics that are absent in the target class, while excluding ambiguous classes. To this end, we first devise innovative strategies to adaptively select high-quality positive and negative classes from the label space, by modeling both the confidence and rank of a class in relation to those of the target class. Then, we introduce novel positive and negative losses designed to guide the learning process, pushing predictions closer to positive classes and away from negative classes. Finally, the positive and negative processes are integrated into a hybrid positive-negative learning framework, facilitating the utilization of non-target classes in both labeled and unlabeled videos. Experimental results on THUMOS14 and ActivityNet v1.3 demonstrate the superiority of the proposed method over prior state-of-the-art approaches.
\end{abstract}

\section{Introduction}
\label{sec:intro}

Temporal Action Localization (TAL) attempts to temporally locate  and recognize action instances of interest in untrimmed videos. It is a fundamental yet challenging task in computer vision, with a wide range of applications, such as security surveillance~\cite{crasto2019mars, yang2020temporal} and human behavior analysis~\cite{diba2018spatio, song2019tacnet}. Traditional TAL approaches heavily rely on large-scale, well-annotated datasets, a process that is both tedious and time-consuming in practice. In response to these challenges, recent efforts have been directed toward Semi-Supervised Temporal Action Localization (SS-TAL), aiming to train models using only a limited number of labeled samples and a substantial amount of unlabeled data.

\begin{figure}[!t]
	\centering
	\includegraphics[width=0.8\linewidth]{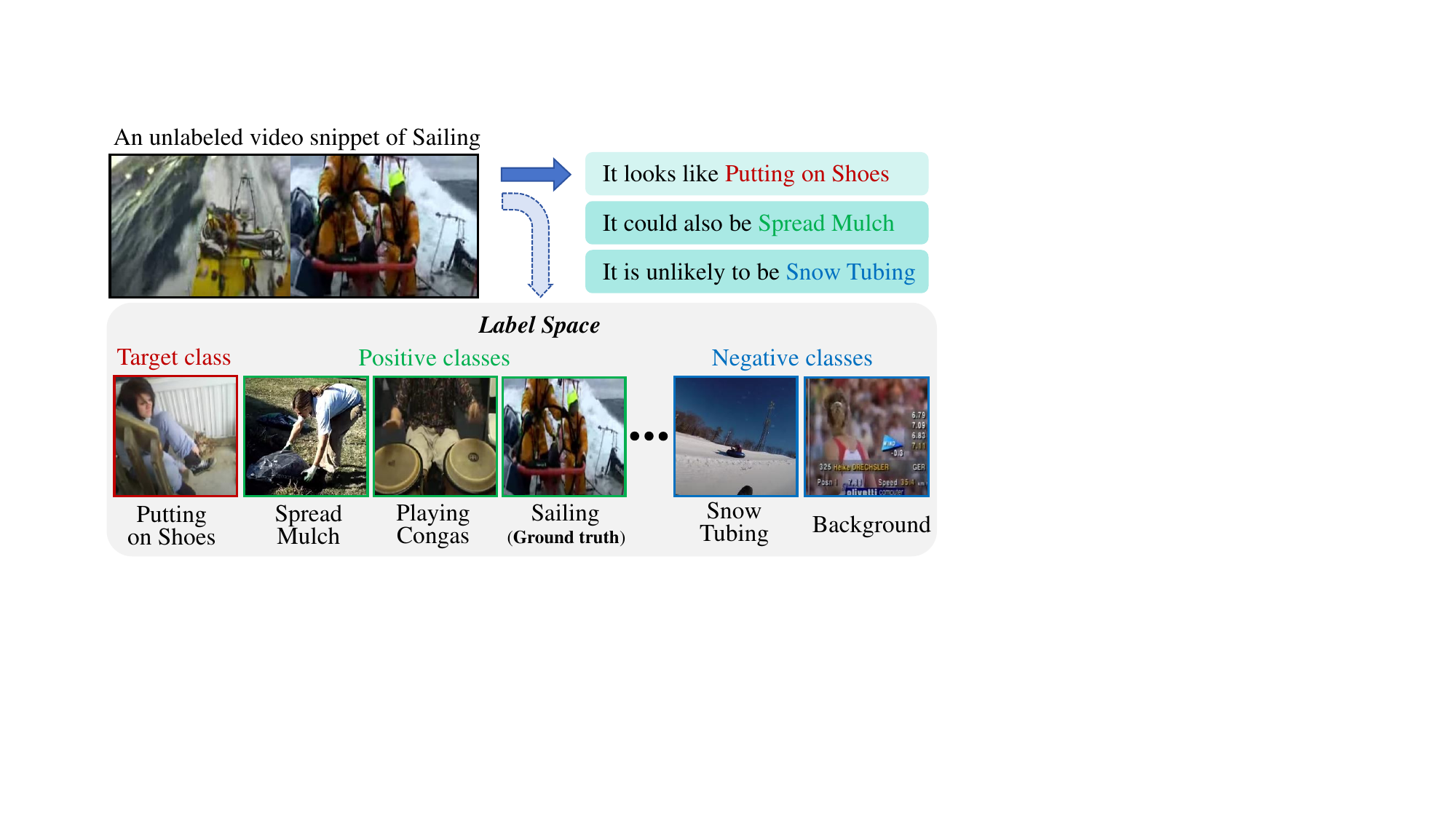}
	\caption{Illustration of unreliable predictions on an unlabeled video snippet. A common practice is to treat the action class with the highest confidence~, \textit{i.e.}, ``Putting on Shoes'' as its target class for model optimization, while the ground truth label~, \textit{i.e.}, ``Sailing'' is buried in the non-target classes.}
	\label{fig:prediction}
\end{figure}

Recent advancements of SS-TAL~\cite{ji2019learning, wang2021self, nag2022semi, xia2023learning} have demonstrated notable success, leveraging two well-known semi-supervised learning paradigms: consistency regularization and self-training.  
Consistency regularization approaches~\cite{ji2019learning, wang2021self} aim to generate reliable predictions through a teacher model to guide the learning process of the student model. However, learning a decent teacher model with limited labeled data is as challenging as the SS-TAL task itself. More recently, self-training approaches~\cite{nag2022semi, xia2023learning} tailored for SS-TAL have dominated this area, attaining state-of-the-art performance. These approaches iteratively use the current model to assign pseudo labels to unlabeled videos and train a new model on both the labeled videos and the pseudo-labeled videos.

Despite achieving promising results, existing approaches simply utilize the \textit{target class} (\ie, the predicted class with the highest confidence) as the pseudo label, which has two significant drawbacks. First, the target class tends to be highly noisy, given that the model is trained on a limited amount of labeled data. Second, the \textit{non-target classes} are entirely disregarded, even though they often contain valuable cues about the action. An illustrative example is depicted in Figure~\ref{fig:prediction}. A video snippet of "Sailing" is mistakenly assigned the target class "Putting on Shoes" for self-training, leading to noisy pseudo labels, while the semantics of the ground truth label are buried among the ignored non-target classes.

In this paper, we approach Semi-Supervised Temporal Action Localization from a novel perspective by learning informative semantics from non-target classes, moving beyond the traditional focus on the target class. Given a predicted class probability distribution on unlabeled data, we often observe two phenomena. First, when the ground truth label does not align with the target class, it frequently falls within other top-ranked classes in the prediction. Second, it is highly unlikely that the low-confidence or bottom-ranked classes contain the ground truth label.

Building upon this observation, we partition the \textit{label space} of the predicted class probability distribution into four subspaces: \textit{target class}, \textit{positive classes}, \textit{negative classes}, and \textit{ambiguous classes}. As mentioned earlier, the target class is defined as the highest-confidence class. Positive classes encompass non-target classes with high confidences, often covering the ground truth class. Negative classes comprise non-target classes with low confidences, making them unlikely to contain the ground truth class. The remaining non-target classes form the ambiguous classes.

While the idea of learning from non-target classes is intriguing, two key challenges need to be addressed: \textit{How should the non-target classes, especially the positive and negative classes, be identified from the predicted class distribution?} \textit{How can the model effectively learn from these non-target classes?} In response to the first challenge, we devise innovative strategies to \textit{adaptively} select high-quality positive and negative classes from the label space. This involves modeling both the confidence and rank of a class in relation to those of the target class. To tackle the second challenge, we introduce novel positive and negative losses designed to push the prediction closer to the positive classes and push it away from the negative classes. Consequently, \textit{positive learning} empowers the model to extract richer semantics relevant to the true class but absent in the target class, while \textit{negative learning} reinforces the model's belief of which classes are incorrect. Given the high uncertainty and noise associated with ambiguous classes, we exclude them from the training process.
Finally, we integrate the positive and negative learning processes into a hybrid positive-negative learning framework to leverage the non-target classes across both labeled and unlabeled videos.

The main contributions of this paper are summarized as follows:
\begin{itemize}
\item This paper introduces a novel paradigm for SS-TAL by emphasizing learning from non-target classes, transcending the conventional focus solely on the target class. The approach involves partitioning the label space of the predicted class distribution into different subspaces, aiming to mine both positive and negative semantics that are absent in the target class, while excluding ambiguous classes.

\item Key aspects of this novel paradigm include identifying the positive and negative classes and learning from these non-target classes. The paper introduces innovative strategies for adaptively selecting high-quality positive and negative classes from the label space. Additionally, new positive and negative losses are proposed to guide the non-target learning effectively. These processes are integrated into a hybrid positive-negative learning framework, facilitating the utilization of non-target classes in both labeled and unlabeled videos.
\item We evaluate the proposed approach on THUMOS14 and ActivityNet v1.3 under a wide range of training settings. Extensive experiments demonstrate that our approach surpasses the previous state-of-the-art methods.
\end{itemize}

\section{Related Work}
\label{sec:related}
\textbf{Fully-Supervised Temporal Action Localization} has witnessed significant advancements in recent years through using plentiful well-annotated videos. Concretely, early \textit{anchor-based} methods~\cite{chao2018rethinking, xu2020g, wang2022rcl} typically employ the multi-scale anchors and attach a classification head and a boundary regression head to refine these pre-defined anchors.
\textit{Anchor-free} methods~\cite{lin2018bsn, zhao2020bottom, lin2021learning, xia2022learning, nag2022temporal} directly regress the boundary locations or perform frame-level action classification to reduce the complexity.
Current prevailing \textit{Transformer-based} methods~\cite{zhang2022actionformer, liu2022end, shi2022react, kim2023self} tackle temporal action localization in a Transformer encoder-decoder framework, which models action instances as a set of learnable action queries. 

\noindent\textbf{Semi-Supervised Temporal Action Localization} leverages valuable information from the unlabeled data with lower annotation cost.  
Existing arts~\cite{ji2019learning, wang2021self, ding2021kfc, nag2022semi, xia2023learning} benefit from the development of general \textit{semi-supervised learning}~\cite{sohn2020fixmatch, tarvainen2017mean} and follow two frameworks, \textit{i.e.}, consistency regularization and self-training.
Ji~\textit{et al.}~\cite{ji2019learning} design two essential types
of sequential perturbations to make consistent action proposal predictions for both teacher and student models. Nag~\textit{et al.}~\cite{nag2022semi} develop a proposal-free temporal masking model to solve the localization error propagation problem. Xia~\textit{et al.}~\cite{xia2023learning} tackle the label noise problem and present a noise-tolerant framework to update the model with reliable pseudo labels that are strictly screened. 
\begin{figure*}[!t]
	\centering
	\includegraphics[width=1\linewidth]{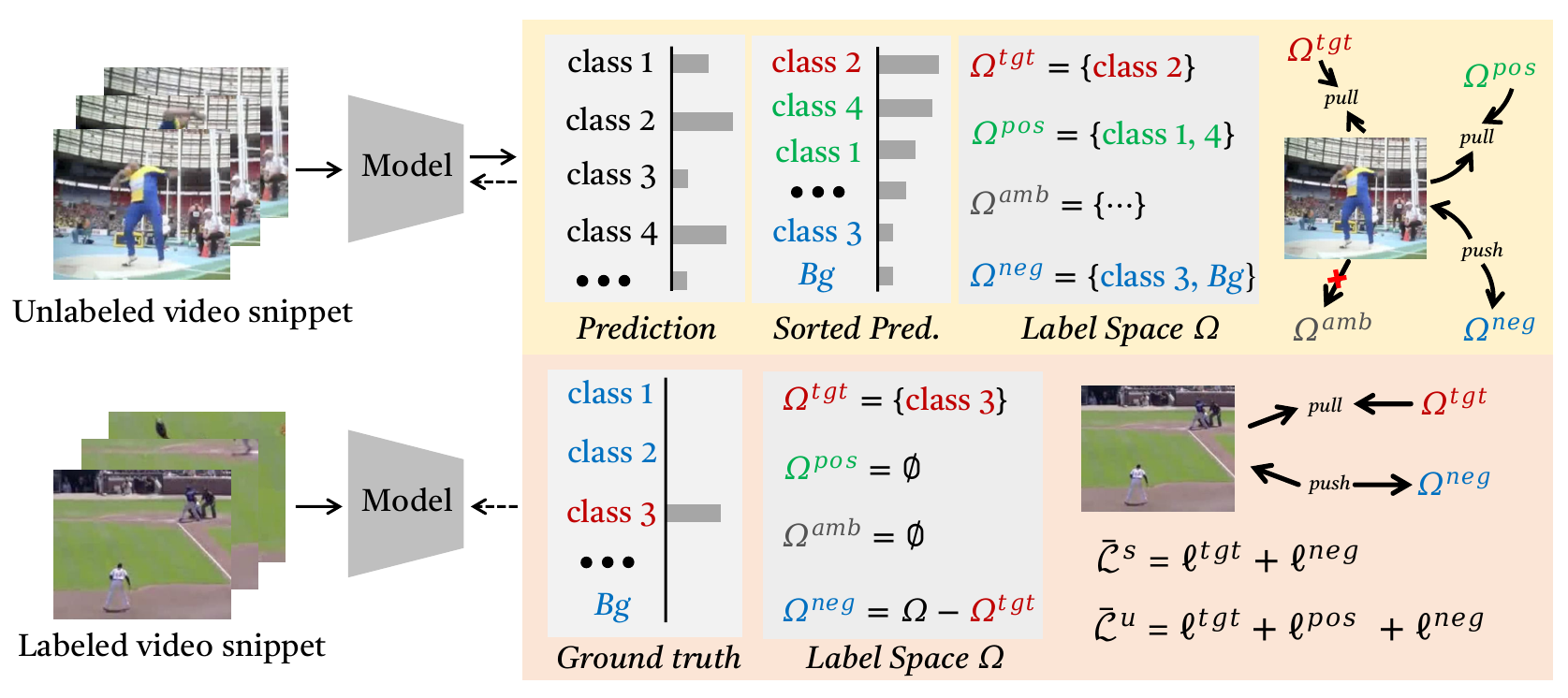}
	\caption{\textbf{An overview of our proposed Non-target Classes Learning framework}. It follows the self-training paradigm, which iteratively uses the
current model to assign pseudo labels to unlabeled videos and trains a new model on both the labeled videos and the pseudo-labeled videos. Given an unlabeled video snippet, the current model predicts a probability distribution of all classes. Our method adaptively partitions the label space $\Omega$ into a target class $\Omega^{tgt}$, positive classes $\Omega^{pos}$, negative classes $\Omega^{neg}$, and ambiguous classes $\Omega^{amb}$, by modeling both the confidence and rank of a class in relation to those of the target class. Based on the label space partition, we design the new positive learning loss $\ell_{pos}$ and negative learning loss $\ell_{neg}$ to mine positive and negative semantics that are absent in the target class, while excluding ambiguous classes.}
	\label{fig:overview}
\end{figure*}

\noindent\textbf{Learning on Pseudo Labels} 
is an important yet key technology in semi-supervised learning. However, most approaches~\cite{chen2022label, zhou2021instant, li2019learning, jin2022semi, qiao2023fuzzy} are limited to learning directly from the target class, so it is inevitable that the model will be misled by noisy pseudo labels. 
Chen~\textit{et al.}~\cite{chen2022label} present a proposal self-assignment for pseudo label assignment, which injects the proposals from student into teacher and generates accurate pseudo labels to match each proposal in the student model accordingly.
Apart from above methods, the complementary label has been used to specify a class that a sample does not belong to~\cite{ishida2017learning}. 
Yu~\textit{et al.}~\cite{yu2018learning} theoretically analyze the problem of biased complementary labels and propose to estimate transition probabilities with no bias. Kim~\textit{et al.}~\cite{kim2019nlnl} aim at learning clean data with ground truth labels while training noise data with a randomly selected label as a complementary label. 

Differing from existing methods, we introduce a novel negative learning approach that adaptively selects richer negative classes based on the confidence of the target class. These negative classes are more informative, reducing the risk of selecting the true label. Additionally, our new positive learning method extracts additional semantics relevant to the true class that may be absent in the target class.

\section{Method}
\label{sec:method}

\subsection{Preliminaries}
\label{sec:pre}
\textbf{Problem Setting}. Given a smaller set of $N^l$ labeled videos $\{X_i^{l}, Y_i^{l}\}_{i=1}^{N^l}$ and a larger set of $N^u$ unlabeled videos $\{X_i^{u}\}_{i=1}^{N^u}$, semi-supervised temporal action localization (SS-TAL) aims to improve action detection by effectively learning from both labeled and unlabeled data. The annotation $Y_i^{l}$ of each labeled video contains the start time, end time, and action category of each action instance. 

\noindent\textbf{Feature Embedding}. For a video $X$, following conventions~\cite{wang2017untrimmednets, nag2022temporal}, we extract its snippet-level features $\{\boldsymbol{x}_i\}_{i=1}^{N^v}$ from consecutive frames by a fine-tuned two-stream network, where $N^v$ is the number of video snippets.

\noindent\textbf{Baseline Model}. Recent works~\cite{nag2022semi, xia2023learning} formulate SS-TAL as a snippet-level classification task. Our method also adopts the proposal-free framework with self-training for SS-TAL, which locates action instances by a classification head and a mask head, as in prior arts~\cite{nag2022semi, xia2023learning}. The learning objective is to minimize the loss function below: 
\begin{equation}
	\mathcal{L}=\mathcal{L}^{s}+\alpha \mathcal{L}^{u},
\end{equation}
where $\mathcal{L}^{s}$ and $\mathcal{L}^{u}$ denote the supervised loss and the unsupervised loss applied on labeled videos and unlabeled videos, respectively, and $\alpha$ is a hyper-parameter. 
The main purpose of the action detection model is to learn the parameters $\theta$ of a model $\mathbb{F}\left(\cdot; \theta\right)$ by optimizing a cross-entropy (CE) loss function on both labeled and unlabeled data:
\begin{equation}
		\mathcal{L}^s =\frac{1}{N^v} \sum_{i=1}^{N^v} \ell^{c e}\left(\mathbb{F}\left(\boldsymbol{x}_i^l ; \theta\right), \boldsymbol{y}_i^l\right),
	\end{equation}
	\begin{equation}
		\mathcal{L}^u =\frac{1}{N^v} \sum_{i=1}^{N^v} \ell^{c e}\left(\mathbb{F}\left(\boldsymbol{x}_i^u ; \theta\right), \boldsymbol{y}_i^u\right),
\end{equation}
where $\boldsymbol{x}_i^l$ and $\boldsymbol{x}_i^u$ are respectively the $i$-th snippet feature vectors of a labeled video and an unlabeled video. $\boldsymbol{y}_i^l \in \mathbb{R}^{C+1}$  and $\boldsymbol{y}_i^u \in \mathbb{R}^{C+1}$ are the one-hot vectors of their ground truth label and pseudo label, respectively, including $C$ action classes and a background class.

\subsection{Motivation}
\label{sec:motivation}
Existing approaches simply utilize the \textit{target class} (\ie, the predicted class with the highest confidence) as the pseudo label. The target class tends to be highly noisy, given that the model is trained on a limited amount of labeled data, thereby significantly degrading the self-training. This paper moves beyond the traditional focus on the target class and addresses SS-TAL from a novel perspective, by learning informative semantics from non-target classes. The motivation for our approach stems from two key observations regarding a predicted class probability distribution on unlabeled data.  First, when the ground truth label does not align with the target class, it frequently falls within other top-ranked classes in the prediction. Second, it is highly unlikely that the low-confidence or bottom-ranked classes contain the ground truth label.

Building upon these observations, we divide the label space of the predicted class probability distribution on an unlabeled video snippet into four subspaces:
\begin{equation}
	\Omega = \left\{1, \ldots, C+1\right\}= \Omega^{tgt} \cup \Omega^{pos} \cup \Omega^{neg} \cup \Omega^{amb},
\end{equation}
where $\Omega^{tgt}$ only holds the target class while $\Omega^{pos}$, $\Omega^{neg}$ and $\Omega^{amb}$ are the positive classes, negative classes, and ambiguous classes, respectively. Positive classes encompass non-target classes with high confidences, often covering the ground truth class. Negative classes comprise non-target classes with low confidences, making them unlikely to contain the ground truth class. The remaining non-target classes form the ambiguous classes. Complementary to traditional target class-based learning (Sec. \ref{sec:tgt}),  \textit{negative learning} (Sec. \ref{sec:neg}) reinforces the model's belief of which classes are incorrect, while \textit{positive learning} (Sec. \ref{sec:pos}) empowers the model to extract richer semantics relevant to the true class but absent in the target class. Given the high uncertainty and noise associated with ambiguous classes, we exclude them from self-training.

\subsection{Learning from Target Class}
\label{sec:tgt}
Existing approaches first obtain the probability distribution $\boldsymbol{p} =  \mathbb{F}\left(\boldsymbol{x}^u ; \theta\right)$ from an unlabeled snippet $\boldsymbol{x}^u$, and then use $argmax_c (p_c)$ as its target class to construct the one-hot pseudo label vector $\boldsymbol{y}^{u}$. The learning objective is formulated as the cross-entropy loss between the model prediction and the target class: 
\begin{equation}
	\ell^{tgt}=-\sum_{c=1}^{C+1} y^u_c \log p_c,
\end{equation}
where $C$ is the number of action classes and $y^u_c \in \{0, 1\}$ represents whether the target class is present.
The model is trained by maximizing the log-likelihood of the target class.
\subsection{Learning from Negative Classes}
\label{sec:neg}
As previously discussed in Sec.~\ref{sec:motivation}, the model may exhibit uncertainty regarding whether a video snippet belongs to the noisy target class but can be fairly certain that it does not belong to negative classes. To effectively learn negative information, a negative class is chosen from non-target classes, and the model is then trained using a negative learning loss given by:
\begin{equation}
	\tilde{\ell}^{neg}=-\log \left(1-p_{c^{neg}}\right),
\end{equation}
which aims to minimize the log-likelihood on the negative class. However, selecting suitable negative classes is challenging.  On the one hand, only selecting one negative class is insufficient to learn valuable negative information. On the other hand, regarding all non-target classes as negative classes would carry the risk of negatively learning the ground truth semantics buried in non-target classes. Therefore, we design an adaptive negative learning strategy to tackle this challenge.

Specifically, let $\boldsymbol{p}=\left[p_{1}, \ldots, p_{C+1}\right]$ denote the class probability distribution predicted on an unlabeled video snippet. Then, we sort it in ascending order of the confidence:
\begin{equation}
    \hat{\boldsymbol{p}} = 
	sorted(\boldsymbol{p})=\left[min(\boldsymbol{p}), \ldots, max(\boldsymbol{p})\right],
\end{equation}
where $max(\boldsymbol{p})$ corresponds to the confidence of the target class. The higher $max(\boldsymbol{p})$ is, the more certain the model is that the target class aligns with the ground truth class.
It also means that we can treat more non-target classes as negative classes for learning negative information.
This line of reasoning motivates us to design an adaptive negative learning strategy by taking the confidence of the target class as reference.
Concretely, we first compute the cumulative probability of its bottom-$k$ classes. If the cumulative probability is less than $max(\boldsymbol{p})$, these $k$ classes will be treated as negative classes that contribute equivalently to negative learning, which could be formulated as:
\begin{equation}
	\Omega^{neg} = \left\{k : \sum_{c=1}^{k} \hat{p}_c \leqslant max(\boldsymbol{p})\right\},
\end{equation}
where $\Omega^{neg}$ holds $k$ negative classes that meet the above criteria. When $max(\boldsymbol{p})$ is very high, it suggests that low-confidence classes carry a lower risk of containing ground truth semantics; therefore, the model will involve more low-confidence classes into $\Omega^{neg}$ for negative learning. When $max(\boldsymbol{p})$ is very low, the ground truth semantics will be more likely to be buried in low-confidence classes; therefore, the model will only select a few negative classes since the cumulative probability of bottom-$k$ classes is small. Based on the negative classes, we reformulate the negative learning loss as:
\begin{equation}
	\ell^{neg}=-\sum_{c\in \Omega^{neg}} \log \left(1-p_c\right).
	\label{eq:nl}
\end{equation}
Our proposed adaptive negative learning will enable the model to effectively learn underlying negative information from as many negative classes as possible. 

\subsection{Learning from Positive Classes}
\label{sec:pos}
Learning from the remaining non-target classes (excluding negative classes) is intriguing as the ground truth semantics are buried among them.  However, learning positive information from all remaining non-target classes is suboptimal since ambiguous classes would confuse the model.  Therefore, we leverage the confidence of the target class as an informative indicator to select positive classes:
\begin{equation}
	\Omega^{pos} = \left\{k : \hat{p}_k \geqslant \lambda  \cdot max(\boldsymbol{p})\right\},
\end{equation}
where $\Omega^{pos}$ holds $k$ positive classes that meet the above criteria and $\lambda$ is a hyper-parameter. In this way, the model will only select the classes whose confidences are close to the target class since they are likely to share similar information related to the ground truth class. Based on the positive classes, we formulate the positive learning loss as:
\begin{equation}
	\ell^{pos}=-\sum_{c \in \Omega^{pos}} \log p_c.
\end{equation}
The positive learning empowers the model to extract richer semantics relevant to the true class but absent in the target class.
\subsection{Hybrid Positive-Negative Learning}
\label{sec:disc}
Finally, we integrate the proposed negative learning and positive learning into our semi-supervised TAL framework.
In training, for all labeled data, the ground truth labels are treated as the target classes with no doubt. The remaining classes, \textit{i.e.}, all non-target classes, will act as negative classes for negative learning, since they are completely unrelated to the ground truth label. Thus, we apply the cross-entropy loss and negative loss for all labeled data. For unlabeled data, we apply the cross-entropy loss for target classes, the positive and negative losses for positive and negative classes as mentioned above, respectively. The overall loss function is shown below:
\begin{equation}
	\mathcal{L}= \bar{\mathcal{L}}^s + \bar{\mathcal{L}}^u
	+ \mathcal{L}^m + \mathcal{L}^{ref} + \mathcal{L}^{rec},
\end{equation}
where the supervised loss $\bar{\mathcal{L}}^s$ contains the cross-entropy loss $\ell^{tgt}$ and the negative learning loss $\ell^{neg}$. The unsupervised loss $\bar{\mathcal{L}}^u$ contains the positive learning loss $\ell^{pos}$ and the negative learning loss $\ell^{neg}$ as well as $\ell^{tgt}$. In addition, the SS-TAL model is mainly composed of a classification head and a mask head, which is further optimized by the mask learning loss $\mathcal{L}^m$, the refinement loss $\mathcal{L}^{ref}$, and the feature reconstruction loss $\mathcal{L}^{rec}$, as in \cite{nag2022semi, xia2023learning}.

In inference, the model generates action instance predictions for each testing video by the classification and mask predictions, as in SPOT~\cite{nag2022semi}. 
More specifically, we can obtain candidate snippets by using a classification threshold and a localization threshold on the classification and mask heads, respectively. Therefore, only the video snippets with high class probabilities and mask scores are selected as top scoring snippets. We use a set of thresholds to produce sufficient candidates. 
For each candidate, we compute its confidence score by multiplying the classification probability and mask score. In post-processing, Soft-NMS~\cite{bodla2017soft} is finally applied to obtain top scoring results.

\section{Experiments}
\label{sec:experiment}
\begin{table*}[!t]
	\centering
        \caption{Main results on THUMOS14 and ActivityNet v1.3 with different percentages of labeled videos, where Baseline refer to the baseline model without positive and negative learning losses. Notably, SSP and SSTAP employ UntrimmedNet~\cite{wang2017untrimmednets} trained with 100\% class labels for proposal classification. }
	\label{tab:comparisonsOnTHUMOSAnet}
	\resizebox{1.0\linewidth}{!}{
		\setlength{\tabcolsep}{0.4em}%
		\begin{tabular}{c|c|c|ccccc|c|ccc|c}
			\toprule
			\specialrule{0em}{1.2pt}{1.2pt}
			\multirow{2}{*}{Label} & \multirow{2}{*}{Method} & \multirow{2}{*}{Backbone} & \multicolumn{6}{c}{THUMOS14~(\%)} & \multicolumn{4}{|c}{ActivityNet v1.3~(\%)} \\
			\cmidrule{4-13}
			& & & 0.3 & 0.4 & 0.5 & 0.6 & 0.7 & Avg. & 0.5 & 0.75 & 0.95 & Avg.\\
			\midrule
			\multirow{6}{*}{10\%}
			& SSP~\cite{ji2019learning} & TSN & 44.2 & 34.1 & 24.6 & 16.9 & 9.3 & 25.8 & 38.9 & 28.7 & 8.4 & 27.6 \\
			& SSTAP~\cite{wang2021self} & TSN & 45.6 & 35.2 & 26.3 & 17.5 & 10.7 & 27.0 & 40.7 & 29.6 & 9.0 & 28.2 \\
			& SPOT~\cite{nag2022semi} & TSN & 49.4 & 40.4 & 31.5 & 22.9 & 12.4 & 31.3 & 49.9 & 31.1 & 8.3 & 32.1 \\
			& NPL~\cite{xia2023learning} & TSN & 50.0 & 41.7 & 33.5 & 23.6 & 13.4 & 32.4 & 50.9 & 32.0 & 7.9 & 32.6 \\
			& Baseline & TSN & 49.2 & 40.2 & 32.8 & 22.4 & 12.5  & 31.4 & 51.2 & 31.8 & 7.3 & 32.1\\
			& \textbf{Ours} & TSN & \textbf{50.9} & \textbf{42.3} & \textbf{34.9} & \textbf{24.7} & \textbf{14.6} & \textbf{33.5} & \textbf{53.0}	& \textbf{34.4}	& \textbf{9.2} & \textbf{34.5}\\
			\midrule
			\multirow{4}{*}{20\%}
			& SPOT~\cite{nag2022semi} & TSN & 52.6 & 43.9 & 34.1 & 25.2 & 16.2 & 34.4 & 51.7 & 32.0 & 6.9 & 32.3 \\
			& NPL~\cite{xia2023learning} & TSN & 53.9 & 45.6 & 36.2 & 26.9 & 16.5 & 35.8 & 52.1 & 32.9 & 7.9 & 32.9 \\
			& Baseline & TSN & 52.8 & 44.0 & 34.2 & 25.4 & 16.0 & 34.5 & 51.8 & 32.2 & 7.0 & 32.4 \\
			& \textbf{Ours} & TSN & \textbf{54.6} &\textbf{46.4} & \textbf{37.0} & \textbf{27.2} & \textbf{17.1} & \textbf{36.5} & \textbf{53.5}	& \textbf{34.7}	& \textbf{9.4} & \textbf{34.8}\\
			\midrule
			\multirow{4}{*}{40\%}
			& SPOT~\cite{nag2022semi} & TSN & 54.4 & 45.8 & 37.2 & 29.7 & 19.4 & 37.3 & 53.3 & 33.0 & 6.6 & 33.2 \\
			& NPL~\cite{xia2023learning} & TSN & 56.2 & 46.7 & 38.8 & 30.3 & 19.5 & 38.3 & 53.4 & 33.9 & 8.1 & 33.8 \\
			& Baseline & TSN & 54.8 & 45.9 & 37.3 & 29.9 & 19.1 & 37.4 & 53.5 & 33.2 & 6.9 & 33.4\\
			& \textbf{Ours} & TSN & \textbf{57.5} & \textbf{48.0} & \textbf{39.6} & \textbf{31.5} & \textbf{21.4} & \textbf{39.6} & \textbf{54.1}	& \textbf{35.6}	& \textbf{9.4} & \textbf{35.4}\\
			\midrule
			\multirow{6}{*}{60\%}
			& SSP~\cite{ji2019learning} & TSN & 53.2 & 46.8 & 39.3 & 29.7 & 19.8 & 37.8 & 49.8 & 34.5 & 7.0 & 33.5 \\
			& SSTAP~\cite{wang2021self} & TSN & 56.4 & 49.5 & 41.0 & 30.9 & 21.6 & 39.9 & 50.1 & 34.9 & 7.4 & 34.0 \\
			& SPOT~\cite{nag2022semi} & TSN & 58.9 & 50.1 & 42.3 & 33.5 & 22.9 & 41.5 & 52.8 & 35.0 & 8.1 & 35.2 \\
			& NPL~\cite{xia2023learning} & TSN & 59.0 & 51.4 & 42.9 & 34.3 & 23.3 & 42.2 & 53.9 & 35.8 & 8.5 & 35.7\\
			& Baseline & TSN & 58.7 & 50.0 & 42.6 & 33.7 & 23.0 & 41.6 & 52.9 & 34.9 & 7.9 & 35.0 \\
			& \textbf{Ours} & TSN & \textbf{59.9} & \textbf{52.6} & \textbf{43.9} & \textbf{35.7} & \textbf{24.0} & \textbf{43.2} & \textbf{54.4}	& \textbf{35.8}	& \textbf{9.5} & \textbf{35.9}\\
			\bottomrule
		\end{tabular}
	}
\end{table*}

\subsection{Datasets and Metrics}
\textbf{Evaluation Datasets.} Following conventions~\cite{zhang2022actionformer, nag2022temporal}, we evaluate our proposed method on two challenging TAL benchmarks, \textit{i.e.}, THUMOS14~\cite{jiang2014thumos} and ActivityNet v1.3~\cite{caba2015activitynet}.
THUMOS14~\cite{jiang2014thumos} contains 200 validation videos and 213 testing videos, including 20 action categories. It is very challenging since each video has more than 15 action instances. Following the common setting~\cite{zeng2021graph}, we use the validation set for training and evaluate on the testing set. 
ActivityNet v1.3~\cite{caba2015activitynet}
is a large-scale benchmark for video-based action localization. It contains 10k training videos and 5k validation videos corresponding to 200 different actions. Following the standard practice~\cite{liu2021multi}, we train our method on the training set and test it on the validation set.

\noindent\textbf{Evaluation Metrics.}
We use the mean Average Precision~(mAP) as the evaluation metric. The tIoU thresholds are $[0.3:0.1:0.7]$ for THUMOS14 and $[0.5:0.05:0.95]$ for ActivityNet v1.3. We report the average mAP of the IoU thresholds between 0.5 and 0.95 with the step of 0.05 on ActivityNet v1.3. Also, we present the average mAP of the tIoU thresholds from 0.3 to 0.7 on THUMOS14.
\subsection{Implementation Details}
Following the conventional setting~\cite{wang2021self, kim2023self, nag2022semi}, we extract each video snippet feature over every fixed consecutive frames by TSN~\cite{wang2016temporal} pre-trained on Kinetics~\cite{xiong2016cuhk}. The temporal dimension is fixed at
100 and 256 for ActivityNet v1.3 and THUMOS14, respectively. 
Our action localization framework adopts the popular proposal-free approach SPOT~\cite{nag2022semi}, which is mainly composed of a classification head and a mask head. Our main contributions focus on the classification head, which originally adopts the cross-entropy loss for target classes. Also, we employ another anchor-free approach Actionformer~\cite{zhang2022actionformer} with the I3D backbone~\cite{carreira2017quo} for fair comparisons.

For semi-supervised setting, we first pre-train our model on the training set for 12 epochs and then we fine-tune the pre-trained model for 15 epochs with a learning rate of $10^{-4}$ for ActivityNet v1.3 and $10^{-5}$, and a cosine learning rate decay is used. 
Following SPOT~\cite{nag2022semi}, we adopt the same label sharpening operator and the threshold set for mask. The Soft-NMS~\cite{bodla2017soft} is
performed on ActivityNet v1.3 and THUMOS14 with a threshold of 0.6 and 0.4, respectively. $\alpha = 1$.
For the labeling ratios, we introduce four SS-TAL settings with different label sizes. Following NPL~\cite{xia2023learning}, we randomly select 10\%, 20\%, 40\%, and 60\% training videos as the labeled set and the remaining as the unlabeled set. Both labeled and unlabeled sets are accessible
for SS-TAL model training. 
\subsection{Comparison with State-of-the-art Methods}
The main results are reported in Table~\ref{tab:comparisonsOnTHUMOSAnet}, where we report mAP at different tIoU thresholds and average mAP. We can observe that our method achieves stable performance improvements over previous works across all data splits on both datasets. We also present the performance of our baseline model in Tabel~\ref{tab:comparisonsOnTHUMOSAnet}. We can see that our main contributions achieve significant performance gains, benefiting from the superiority of the proposed framework.

Specifically, for the THUMOS14 dataset, it is a challenging TAL benchmark due to dense action instances and ambiguous semantics. Our method still outperforms all other comparable methods in all labeled ratios, indicating that the performance gains from our positive and negative learning strategies. Especially, our method obtains remarkable performance when the number of labeled data is very limited~(with only 10\% or 20\% labeled videos). It demonstrates that our method could learn underlying valuable information from non-target classes.

The superiority of the proposed method is more emphasized for ActivityNet v1.3, which is a more large-scale video dataset so as to provide a larger label space for effective hybrid positive-negative learning. As depicted in Table~\ref{tab:comparisonsOnTHUMOSAnet}, our method shows a distinct improvement compared to all other methods. Additionally, the improvements suggests that indirectly learning from positive and negative classes further benefits SS-TAL.

\subsection{Ablation Study}

\begin{table}[t]
	\centering
        \caption{Ablation study of different losses on THUMOS14 and ActivityNet v1.3 with 10\% labels, where $\ell^{tgt}$ is the vanilla cross-entropy loss for snippet-level action classification, and $\ell^{neg}$ and $\ell^{pos}$ are the proposed negative and positive learning losses for excavating complementary information.}
	\label{tab:ablation-on-thu}
	\resizebox{0.9\linewidth}{!}{
		\setlength{\tabcolsep}{0.6em}%
		\begin{tabular}{ccc|cccc|cccc}
			\toprule 
			\specialrule{0em}{1.0pt}{1.0pt}
			\multirow{2}{*}{$\ell^{tgt}$} & \multirow{2}{*}{$\ell^{neg}$} & \multirow{2}{*}{$\ell^{pos}$}& \multicolumn{4}{c}{THUMOS14 (\%)} & \multicolumn{4}{|c}{ActivityNet v1.3 (\%)} \\
			\cmidrule{4-11}
			 & & & 0.3  & 0.5  & 0.7 & Avg. & 0.5  & 0.75  & 0.95 & Avg.\\ 
			\midrule 
			\Checkmark &  & & 49.2 &  32.8 &  12.5 &  31.4 & 51.2 & 31.8 & 7.3 & 32.1\\
			\Checkmark  & \Checkmark  & & 50.1 & 34.0 & 13.9 & 32.7 & 52.3 & 32.7  & 8.3 &  33.4 \\
			\Checkmark   & \Checkmark & \Checkmark & \textbf{50.9} & \textbf{34.9} & \textbf{14.6} & \textbf{33.5} & \textbf{53.0} & \textbf{34.4} & \textbf{9.2} & \textbf{34.5}   \\
			\bottomrule 
		\end{tabular}
	}
\end{table}

\noindent\textbf{Effectiveness of loss terms}. 
To prove our core insight, \textit{i.e.}, learning underlying informative semantics from non-target classes, we conduct experiments in Table~\ref{tab:ablation-on-thu} to ablate each loss step by step. Above all, we use $\ell^{tgt}$ trained model as our baseline, achieving average mAP of 31.4\% and 32.1\% on THUMOS14 and Activitynet v1.3, respectively. Applying the proposed $\ell^{neg}$ consistently improves the baseline by a large margin on both benchmarks, arguably since the potential negative information improves the snippet-level semantic discrimination. In addition, the proposed $\ell^{pos}$ also significantly improve the performance of the model by excavating the ground truth label related semantics from the positive classes.

\begin{figure}[t]
	\begin{minipage}[h]{0.5\linewidth}
		\centering
		\captionof{table}{Ablation study of SS-TAL results on THUMOS14 using I3D features and Actionformer~\cite{zhang2022actionformer}, where the label ratio is 10\% and $\star$ represents only using labeled videos.}
		\label{tab:ablation-on-i3d}
		\begin{tabular}{c|c|cccc}
			\toprule 
			\specialrule{0em}{1.0pt}{1.0pt}
			\multirow{2}{*}{Method} & \multirow{2}{*}{Bkb} & \multicolumn{4}{c}{THUMOS14 (\%)} \\
			\cmidrule{3-6}
			& & 0.3  & 0.5  & 0.7 & Avg. \\ 
			\midrule 
			ActF$^{\star}$~\cite{zhang2022actionformer}& I3D & 28.5 & 14.1 & 4.1 & 15.6\\
			NPL~(ActF)~\cite{xia2023learning}& I3D & 32.8 & 20.1 & 7.2 & 20.3  \\
			Ours~(ActF)& I3D & \textbf{34.1} & \textbf{21.3} & \textbf{7.9} & \textbf{21.4}    \\
			\bottomrule 
		\end{tabular}
	\end{minipage}
	\begin{minipage}[h]{0.5\linewidth}
		\centering
            \captionof{table}{Empirical study of hyper-parameter $\lambda$ on THUMOS14 with 10\% labels, where $\lambda$ affects the number of the positive classes for positive learning.}
	    \label{tab:ablation-on-lambda}
		\begin{tabular}{c|cccc}
			\toprule 
			\specialrule{0em}{1.0pt}{1.0pt}
			\multirow{2}{*}{$\lambda$ value} & \multicolumn{4}{c}{THUMOS14 (\%)} \\
			\cmidrule{2-5}
			& 0.3  & 0.5  & 0.7 & Avg. \\ 
			\midrule 
			0.90 & 50.6 & 34.8 &  14.3 & 33.2\\
			0.85 & \textbf{50.9} & \textbf{34.9} & \textbf{14.6} & \textbf{33.5} \\
			0.80 & 50.2 & 34.3 & 14.1 & 33.0  \\
                0.75 & 50.0 & 34.1 & 14.0 & 32.9  \\
			\bottomrule 
		\end{tabular}
	\end{minipage}
\end{figure}
\noindent\textbf{Different backbone and detector}. 
The proposed method features a hybrid positive-negative learning loss to learn complementary information from the label space of each video snippet, which is feature-agnostic and model-agnostic. To verify this point, we combine our method with another popular I3D~\cite{carreira2017quo} backbone in TAL and the powerful detector Actionformer~\cite{zhang2022actionformer}, and then the comparable results are shown in Table~\ref{tab:ablation-on-i3d}. The performance gains support our point, confirming that the superiority of our method is feature-agnostic and model-agnostic.

\noindent\textbf{Empirical study of hyper-parameter $\lambda$}.
Ground truth classes are often hidden in positive classes, which is ignored by target-class-based learning methods.
In contrast, we introduce a hyper-parameter $\lambda$ that adaptively selects the number of positive classes based on the confidence of the sample. Then, we conduct an ablation study to vary the value of $\lambda$ and delve into its impact on the performance. 
From Table~\ref{tab:ablation-on-lambda}, it can be observed that higher $\lambda$ choosing fewer positive classes may make it difficult to fully learn the informative semantics via positive learning while lower $\lambda$ choosing more classes as positive classes may carry the risk of involving unreliable ambiguous classes.

\begin{figure}[!t]
	\centering
	\includegraphics[width=0.7\linewidth]{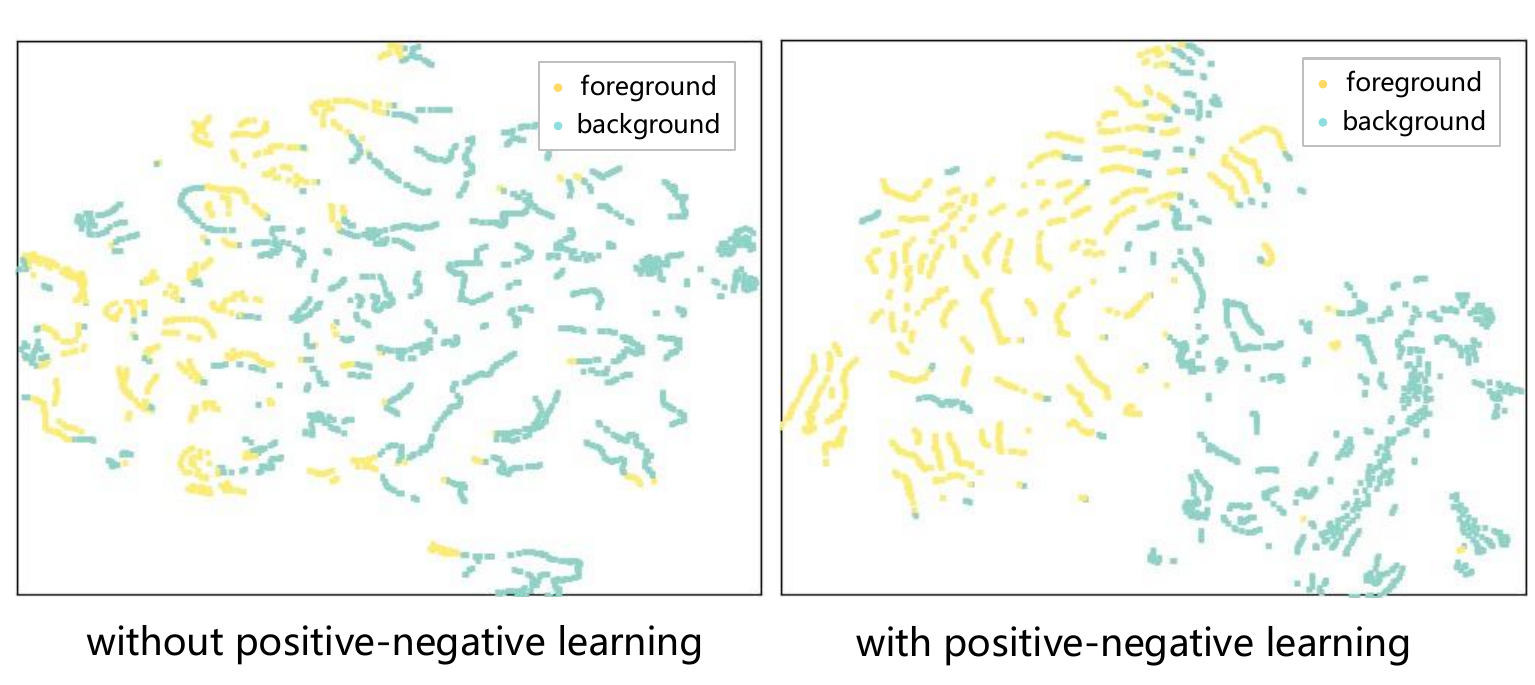}
	\caption{Effect of our method on foreground-background subtask. We present the visualization of foreground feature and background feature on an unlabeled THUMOS14 video.}
	\label{fig:fg-bg}
\end{figure}
\begin{figure}[!t]
	\centering
	\includegraphics[width=0.7\linewidth]{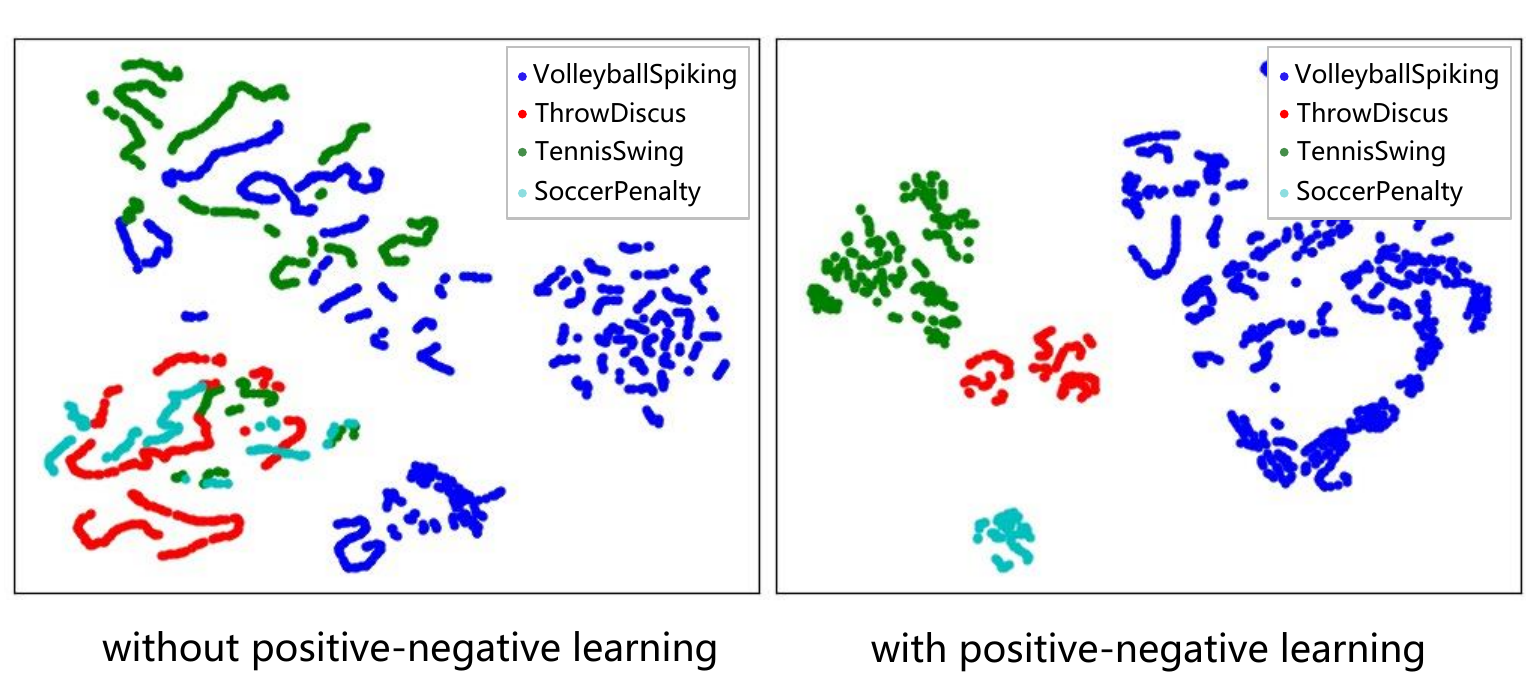}
	\caption{Effect of our method on foreground-instance subtask. We present the visualization of features of four challenging classes on THUMOS14.}
	\label{fig:fg-ins}
\end{figure}

\noindent\textbf{Qualitative analysis of the positive and negative learning}.
Learning complementary information from non-target classes contributes to improving the class-level representation. To verify this point, we present the visualizations of foreground-background features and foreground-instance features in Figure~\ref{fig:fg-bg} and Figure~\ref{fig:fg-ins}, respectively. On the one hand, from Figure~\ref{fig:fg-bg}, the proposed hybrid positive-negative learning separates foreground and background features more clearly by excavating the ground truth semantics hidden in positive classes. On the other hand, from Figure~\ref{fig:fg-ins}, we can observe that the model produces a much clearer boundary of each class with the hybrid positive-negative learning. It shows that our method could improve the generalization ability of the model.

\begin{figure}[!t]
	\begin{minipage}[h]{0.5\linewidth}
		\centering
		\includegraphics[width=0.7\linewidth]{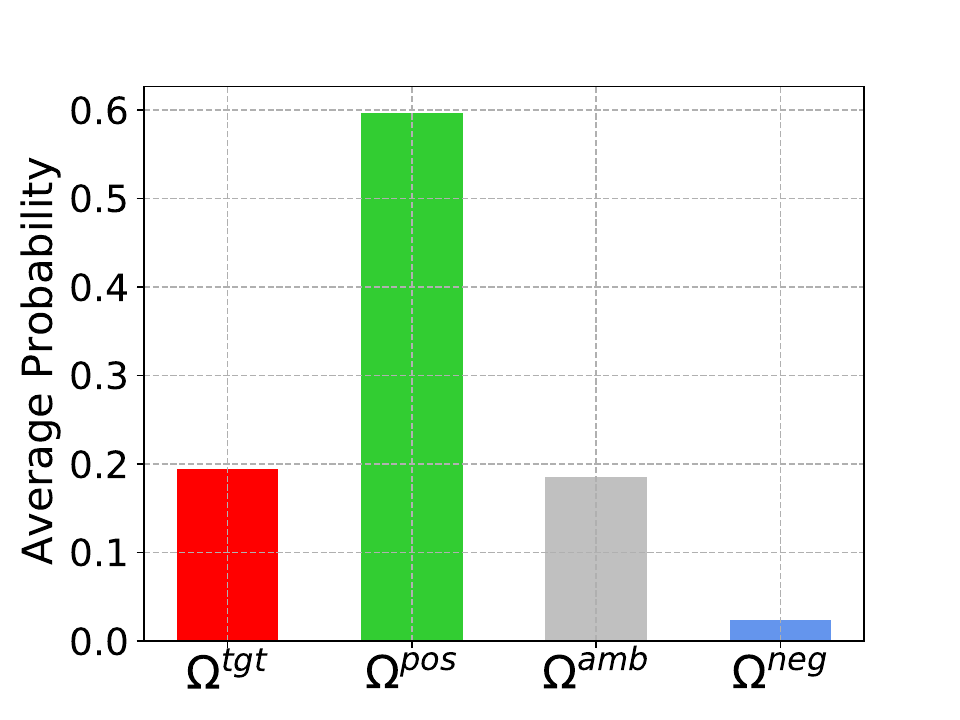}
            \vspace{-0.9em}
            \caption{The average probability that the ground truth label is located in different label subspace, \textit{i.e.} target class $\Omega^{tgt}$, positive classes $\Omega^{pos}$, ambiguous classes $\Omega^{amb}$ and negative classes $\Omega^{neg}$. The experiment is performed on 90\% unlabeled THUMOS14.}
	    \label{fig:prob}
	\end{minipage}
	\begin{minipage}[h]{0.5\linewidth}
		\centering
            \captionof{table}{Comparison with the soft pseudo-label method~\cite{arazo2020pseudo} and complementary label~\cite{ishida2017learning}. The comparison results verify the superiority of our method over previous semi-supervised technologies. }
		\begin{tabular}{c|cccc}
			\toprule 
			\specialrule{0em}{1.0pt}{1.0pt}
			\multirow{2}{*}{Method} & \multicolumn{4}{c}{THUMOS14 (\%)} \\
			\cmidrule{2-5}
			 & 0.3  & 0.5  & 0.7 & Avg. \\ 
			\midrule 
			soft pseudo label & 49.5 & 33.2 & 12.7 & 31.7 \\
			complementary label & 50.0 & 33.5 & 13.1 & 32.1  \\
                Ours & \textbf{50.9} & \textbf{34.9} & \textbf{14.6} & \textbf{33.5}  \\
			\bottomrule 
		\end{tabular}
	\label{tab:ablation-on-methods}
	\end{minipage}
\end{figure}

\noindent\textbf{Quantitative evaluation of label space}.
The proposed method could adaptively divide the entire label space into target class, positive classes, ambiguous classes, and negative classes, and then perform positive and negative learning to excavate ground truth semantics and underlying negative information. The key to performance improvement is whether positive classes contain the ground truth label while negative classes run a low risk of containing the ground truth label.
Thus, we calculate the average probability that the ground truth label is located in these four class subspaces, as shown in Figure~\ref{fig:prob}. It can be observed that our method could effectively use positive classes to mine ground truth semantics and exploit negative classes to improve the model as much as possible.


\noindent\textbf{Comparison with other semi-supervised approaches}. To validate the superiority of our method over previous semi-supervised technologies, we explore the soft pseudo-label method~\cite{arazo2020pseudo} and the complementary label method~\cite{ishida2017learning} (learning from a random non-target class). We incorporate their main ideas into our work. The comparison results in Table~\ref{tab:ablation-on-methods} indicate that the soft pseudo-label produced by the model is quite noisy and includes limited extra knowledge beyond the target label. In contrast to the complementary label, our hybrid positive-negative learning can adaptively extract richer, more informative action semantics from unlabeled videos while reducing the risk of choosing the true label.

\subsection{Visualization results}
\begin{figure}[!t]
	\centering
	\includegraphics[width=1\linewidth]{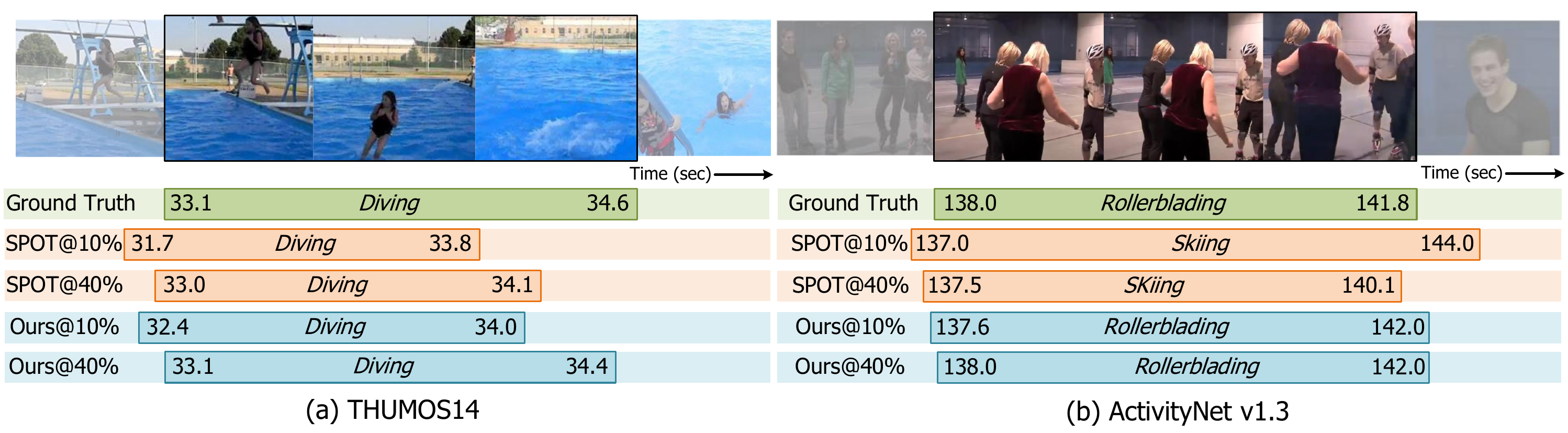}
	\caption{Qualitative SS-TAL result comparison of our proposed method with SPOT~\cite{nag2022semi} on two untrimmed videos from (a) THUMOS14 and (b) ActivityNet v1.3, respectively.}
	\label{fig:vis}
\end{figure}

As shown in Figure~\ref{fig:vis}, we provide some qualitative results by previous work SPOT~\cite{nag2022semi} and our approach, where the model is trained with 10\% and 40\% labeled data on both THUMOS14 and ActivityNet v1.3. 
Benefiting from using non-target classes, our method can locate and recognize the target actions more accurately, demonstrating the superiority of our method. 

\section{Limitation}
This paper proposes to learn informative semantics from non-target classes instead of only target class, benefiting from the abundant information hidden in the label space. Thus, it is difficult for the proposed method to achieve significant performance gains  when the training set and the label space are small. In addition, this work completely excludes all ambiguous classes in training, which may result in some of the indicative information being wasted. Therefore, improving the model by using ambiguous classes will be part of our future work.

\section{Conclusion}
In this paper, we introduce a novel paradigm for SS-TAL by emphasizing learning from non-target classes, transcending the conventional focus solely on the target class. The approach fist partitions the entire label space of the predicted class distribution into different subspaces, aiming to mine both positive and negative semantics that are absent in the target class, while excluding ambiguous classes. Then, we develop innovative strategies for adaptively selecting high-quality positive and negative classes from the label space. Additionally, new positive and negative losses are proposed to guide the non-target learning effectively.
The extensive experiments on two popular benchmarks with consistent performance gains demonstrate the effectiveness of our method.

%
%
\bibliographystyle{splncs04}
\bibliography{eccv}

\begin{thebibliography}{10}
\providecommand{\url}[1]{\texttt{#1}}
\providecommand{\urlprefix}{URL }
\providecommand{\doi}[1]{https://doi.org/#1}

\bibitem{arazo2020pseudo}
Arazo, E., Ortego, D., Albert, P., O’Connor, N.E., McGuinness, K.:
  Pseudo-labeling and confirmation bias in deep semi-supervised learning. In:
  IJCNN. pp.~1--8 (2020)

\bibitem{bodla2017soft}
Bodla, N., Singh, B., Chellappa, R., Davis, L.S.: Soft-nms--improving object
  detection with one line of code. In: ICCV. pp. 5561--5569 (2017)

\bibitem{caba2015activitynet}
Caba~Heilbron, F., Escorcia, V., Ghanem, B., Carlos~Niebles, J.: {ActivityNet}:
  A large-scale video benchmark for human activity understanding. In: CVPR. pp.
  961--970 (2015)

\bibitem{carreira2017quo}
Carreira, J., Zisserman, A.: Quo vadis, action recognition? a new model and the
  kinetics dataset. In: CVPR. pp. 6299--6308 (2017)

\bibitem{chao2018rethinking}
Chao, Y.W., Vijayanarasimhan, S., Seybold, B., Ross, D.A., Deng, J.,
  Sukthankar, R.: Rethinking the faster {R-CNN} architecture for temporal
  action localization. In: CVPR. pp. 1130--1139 (2018)

\bibitem{chen2022label}
Chen, B., Chen, W., Yang, S., Xuan, Y., Song, J., Xie, D., Pu, S., Song, M.,
  Zhuang, Y.: Label matching semi-supervised object detection. In: CVPR. pp.
  14381--14390 (2022)

\bibitem{crasto2019mars}
Crasto, N., Weinzaepfel, P., Alahari, K., Schmid, C.: Mars: Motion-augmented
  rgb stream for action recognition. In: CVPR. pp. 7882--7891 (2019)

\bibitem{diba2018spatio}
Diba, A., Fayyaz, M., Sharma, V., Mahdi~Arzani, M., Yousefzadeh, R., Gall, J.,
  Van~Gool, L.: Spatio-temporal channel correlation networks for action
  classification. In: ECCV. pp. 284--299 (2018)

\bibitem{ding2021kfc}
Ding, X., Wang, N., Gao, X., Li, J., Wang, X., Liu, T.: {KFC}: An efficient
  framework for semi-supervised temporal action localization. IEEE T-IP
  \textbf{30},  6869--6878 (2021)

\bibitem{ishida2017learning}
Ishida, T., Niu, G., Hu, W., Sugiyama, M.: Learning from complementary labels.
  In: NeurIPS (2017)

\bibitem{ji2019learning}
Ji, J., Cao, K., Niebles, J.C.: Learning temporal action proposals with fewer
  labels. In: ICCV. pp. 7073--7082 (2019)

\bibitem{jiang2014thumos}
Jiang, Y.G., Liu, J., Zamir, A.R., Toderici, G., Laptev, I., Shah, M.,
  Sukthankar, R.: {THUMOS} challenge: Action recognition with a large number of
  classes (2014)

\bibitem{jin2022semi}
Jin, Y., Wang, J., Lin, D.: Semi-supervised semantic segmentation via gentle
  teaching assistant. In: NeurIPS. pp. 2803--2816 (2022)

\bibitem{kim2023self}
Kim, J., Lee, M., Heo, J.P.: Self-feedback detr for temporal action detection.
  ICCV  (2023)

\bibitem{kim2019nlnl}
Kim, Y., Yim, J., Yun, J., Kim, J.: Nlnl: Negative learning for noisy labels.
  In: ICCV. pp. 101--110 (2019)

\bibitem{li2019learning}
Li, J., Wong, Y., Zhao, Q., Kankanhalli, M.S.: Learning to learn from noisy
  labeled data. In: CVPR. pp. 5051--5059 (2019)

\bibitem{lin2021learning}
Lin, C., Xu, C., Luo, D., Wang, Y., Tai, Y., Wang, C., Li, J., Huang, F., Fu,
  Y.: Learning salient boundary feature for anchor-free temporal action
  localization. In: CVPR. pp. 3320--3329 (2021)

\bibitem{lin2018bsn}
Lin, T., Zhao, X., Su, H., Wang, C., Yang, M.: {BSN}: Boundary sensitive
  network for temporal action proposal generation. In: ECCV. pp. 3--19 (2018)

\bibitem{liu2021multi}
Liu, X., Hu, Y., Bai, S., Ding, F., Bai, X., Torr, P.H.: Multi-shot temporal
  event localization: a benchmark. In: CVPR. pp. 12596--12606 (2021)

\bibitem{liu2022end}
Liu, X., Wang, Q., Hu, Y., Tang, X., Zhang, S., Bai, S., Bai, X.: End-to-end
  temporal action detection with transformer. IEEE TIP  \textbf{31},
  5427--5441 (2022)

\bibitem{nag2022semi}
Nag, S., Zhu, X., Song, Y.Z., Xiang, T.: Semi-supervised temporal action
  detection with proposal-free masking. In: ECCV (2022)

\bibitem{nag2022temporal}
Nag, S., Zhu, X., Song, Y.Z., Xiang, T.: Temporal action detection with global
  segmentation mask learning. In: ECCV (2022)

\bibitem{qiao2023fuzzy}
Qiao, P., Wei, Z., Wang, Y., Wang, Z., Song, G., Xu, F., Ji, X., Liu, C., Chen,
  J.: Fuzzy positive learning for semi-supervised semantic segmentation. In:
  CVPR. pp. 15465--15474 (2023)

\bibitem{shi2022react}
Shi, D., Zhong, Y., Cao, Q., Zhang, J., Ma, L., Li, J., Tao, D.: React:
  Temporal action detection with relational queries. In: ECCV. pp. 105--121.
  Springer (2022)

\bibitem{sohn2020fixmatch}
Sohn, K., Berthelot, D., Carlini, N., Zhang, Z., Zhang, H., Raffel, C.A.,
  Cubuk, E.D., Kurakin, A., Li, C.L.: {FixMatch}: Simplifying semi-supervised
  learning with consistency and confidence. In: NeurIPS. pp. 596--608 (2020)

\bibitem{song2019tacnet}
Song, L., Zhang, S., Yu, G., Sun, H.: Tacnet: Transition-aware context network
  for spatio-temporal action detection. In: CVPR. pp. 11987--11995 (2019)

\bibitem{tarvainen2017mean}
Tarvainen, A., Valpola, H.: Mean teachers are better role models:
  Weight-averaged consistency targets improve semi-supervised deep learning
  results. In: NeurIPS. pp. 1196--1205 (2017)

\bibitem{wang2017untrimmednets}
Wang, L., Xiong, Y., Lin, D., Van~Gool, L.: Untrimmednets for weakly supervised
  action recognition and detection. In: CVPR. pp. 4325--4334 (2017)

\bibitem{wang2016temporal}
Wang, L., Xiong, Y., Wang, Z., Qiao, Y., Lin, D., Tang, X., Van~Gool, L.:
  Temporal segment networks: Towards good practices for deep action
  recognition. In: ECCV. pp. 20--36 (2016)

\bibitem{wang2022rcl}
Wang, Q., Zhang, Y., Zheng, Y., Pan, P.: {RCL}: Recurrent continuous
  localization for temporal action detection. In: CVPR. pp. 13566--13575 (2022)

\bibitem{wang2021self}
Wang, X., Zhang, S., Qing, Z., Shao, Y., Gao, C., Sang, N.: Self-supervised
  learning for semi-supervised temporal action proposal. In: CVPR. pp.
  1905--1914 (2021)

\bibitem{xia2023learning}
Xia, K., Wang, L., Zhou, S., Hua, G., Tang, W.: Learning from noisy pseudo
  labels for semi-supervised temporal action localization. In: ICCV. pp.
  10160--10169 (2023)

\bibitem{xia2022learning}
Xia, K., Wang, L., Zhou, S., Zheng, N., Tang, W.: Learning to refactor action
  and co-occurrence features for temporal action localization. In: CVPR. pp.
  13874--13883 (2022)

\bibitem{xiong2016cuhk}
Xiong, Y., Wang, L., Wang, Z., Zhang, B., Song, H., Li, W., Lin, D., Qiao, Y.,
  Van~Gool, L., Tang, X.: Cuhk \& ethz \& siat submission to activitynet
  challenge 2016. arXiv preprint arXiv:1608.00797  (2016)

\bibitem{xu2020g}
Xu, M., Zhao, C., Rojas, D.S., Thabet, A., Ghanem, B.: {G-TAD}: Sub-graph
  localization for temporal action detection. In: CVPR. pp. 10156--10165 (2020)

\bibitem{yang2020temporal}
Yang, C., Xu, Y., Shi, J., Dai, B., Zhou, B.: Temporal pyramid network for
  action recognition. In: CVPR. pp. 591--600 (2020)

\bibitem{yu2018learning}
Yu, X., Liu, T., Gong, M., Tao, D.: Learning with biased complementary labels.
  In: ECCV. pp. 68--83 (2018)

\bibitem{zeng2021graph}
Zeng, R., Huang, W., Tan, M., Rong, Y., Zhao, P., Huang, J., Gan, C.: Graph
  convolutional module for temporal action localization in videos. IEEE TPAMI
  \textbf{44}(10),  6209--6223 (2022)

\bibitem{zhang2022actionformer}
Zhang, C., Wu, J., Li, Y.: {ActionFormer}: Localizing moments of actions with
  transformers. In: ECCV. pp. 492--510 (2022)

\bibitem{zhao2020bottom}
Zhao, P., Xie, L., Ju, C., Zhang, Y., Wang, Y., Tian, Q.: Bottom-up temporal
  action localization with mutual regularization. In: ECCV. pp. 539--555 (2020)

\bibitem{zhou2021instant}
Zhou, Q., Yu, C., Wang, Z., Qian, Q., Li, H.: Instant-teaching: An end-to-end
  semi-supervised object detection framework. In: CVPR. pp. 4081--4090 (2021)

\end{thebibliography}
\end{document}